\documentclass{article}

\usepackage{arxiv}

\usepackage[utf8]{inputenc} 
\usepackage[T1]{fontenc}    
\usepackage{hyperref}       
\usepackage{url}            
\usepackage{booktabs}       
\usepackage{amsfonts}       
\usepackage{nicefrac}       
\usepackage{microtype}      
\usepackage{lipsum}		
\usepackage{graphicx}
\usepackage{natbib}
\usepackage{doi}

\usepackage{times}
\usepackage{latexsym}

\usepackage{subcaption}
\usepackage{xstring}
\usepackage{amsmath}
\usepackage{graphicx}
\usepackage{tikz}
\usepackage{tikz-qtree}

\usepackage{microtype}

\title{Causal Transformers Perform Below Chance on Recursive Nested Constructions, Unlike Humans}

\author{\href{https://orcid.org/0000-0001-8774-6427}{\includegraphics[scale=0.06]{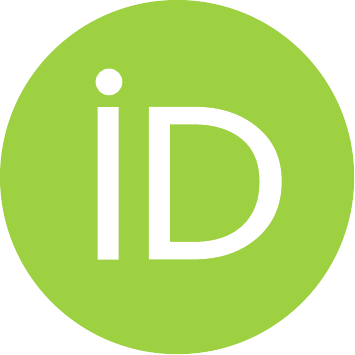}\hspace{1mm}Yair Lakretz} \\
	Cognitive Neuroimaging Unit\\
	NeuroSpin center\\
	91191 Gif-sur-Yvette, France \\
	\texttt{yair.lakretz@gmail.com} \\
	
	\And
	Théo Desbordes \\
	Facebook AI Research\\
	NeuroSpin center \\
	France\\
	\texttt{tdesbordes@fb.com} \\
	
	\And
	Dieuwke Hupkes \\
	Facebook AI Research\\
	Paris, France\\
	\texttt{dieuwkehupkes@fb.com} \\
	
	\And
	Stanislas Dehaene \\
	Cognitive Neuroimaging Unit\\
	NeuroSpin center\\
	Collège de France\\
	Paris, France\\

	}

\renewcommand{\shorttitle}{\textit{arXiv} Template}

\date{}

\begin{document}
\maketitle

\begin{abstract}
Recursive processing is considered a hallmark of human linguistic abilities.
A recent study evaluated recursive processing in recurrent neural language models (RNN-LMs) and showed that such models perform below chance level on embedded dependencies within nested constructions -- a prototypical example of recursion in natural language. Here, we study if state-of-the-art Transformer LMs do any better.
We test four different Transformer LMs on two different types of nested constructions, which differ in whether the embedded (inner) dependency is short or long range.
We find that Transformers achieve near-perfect performance on short-range embedded dependencies, significantly better than previous results reported for RNN-LMs and humans. However, on \emph{long-range} embedded dependencies, Transformers' performance sharply drops below chance level. Remarkably, the addition of only three words to the embedded dependency caused Transformers to fall from near-perfect to below-chance performance. Taken together, our results reveal Transformers' shortcoming when it comes to recursive, structure-based, processing. 

\end{abstract}

\section{introduction}

\begin{figure}
    \centering
    \begin{center}
    \resizebox{0.5\columnwidth}{!}{%
    \begin{tikzpicture}[sibling distance=0.5pt]
    \tikzset{edge from parent/.append style={very thick}}
    \Tree [.S [.NP  [.DP The ] [.NP \textbf{\textcolor{blue}{keys}} [.RC that [ [.NP [.NP the  \textbf{\textcolor{red}{man}} ] [.PP near [.NP the \underline{cabinet} ] ] ] [.VP \textbf{\textcolor{red}{holds}} ] ] ] ] ] [.VP \textbf{\textcolor{blue}{are}} [.PP $..$ ] ] ]
    \end{tikzpicture}
    }
    \end{center}
    \caption{A tree-structure representation of a recursive structure with two long-range dependencies, one nested within the other one.}
    \label{fig:tree}
\end{figure}
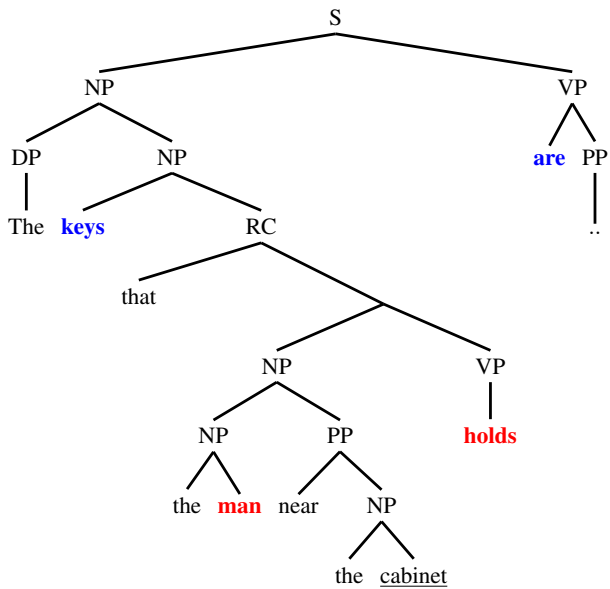

One of the fundamental principles of contemporary linguistics states that language processing requires the ability to deal with nested structures. 
Recursion, a specific type of computation that involves repeatedly applying a function to its own output,
is suggested to be at the core of this ability \citep{Hauser:etal:2002}. 
The strongest evidence for recursion in human language processing arises from the tree-like nested structure of sentences in natural language, in which phrases of a particular type (i.e.,\ NPs) can be embedded in other phrases of that same type (Figure \ref{fig:tree}). Humans, it is argued, are endowed with a unique competence for recursive processing, which allows them to represent and process such nested tree structures \citep{Chomsky:2000, Hauser:etal:2002, Dehaene:etal:2015}.

In recent years, neural language models (NLMs) have shown tremendous advances on a variety of linguistic tasks, such as next-word prediction (also known as "language modeling"), translation or semantic inference.
Furthermore, evaluations of their syntactic abilities have shown promising results, with similar or even above-human performance on a variety of different tasks \citep{marvin-linzen-2018-targeted, goldberg2019assessing,jumelet-etal-2021-language,giulianelli-etal-2018-hood}. However, negative results were recently also presented \citep{warstadt-etal-2020-blimp, hu2020systematic, petty2021transformers}. In particular, when it comes to recursive processing, \citet{lakretz2020nested} showed that while recurrent neural network language models (RNN-LMs) perform well on long-range dependencies, such as the relationship between \textbf{keys} and \textbf{are} in sentences like ``The \textbf{keys} that the \textit{man} near the cabinet \textit{holds}, \textbf{are} red'' (Figure \ref{fig:constructions}b), they perform below chance on the shorter, embedded dependency (\textit{man}-\textit{holds}).
Humans, instead, perform significantly better on such dependencies, although interestingly, for them too, the shorter inner dependency is more difficult than the long outer one. 

The study by \citeauthor{lakretz2020nested} illustrates how investigations of neural networks can inspire experiments about human language processing. However, their study focuses on only a single architecture, an RNN-LM with LSTM units \cite{Hochreiter:Schmidhuber:1997}, which is currently outperformed on many fronts by the newer \emph{Transformer} models \citep{vaswani2017attention}. In this short paper, our main question is therefore whether Transformer models do any better when it comes to processing recursive constructions. We then further explore similarities and differences in performance patterns of RNN and Trasformer language models.

Our main results show that when tested on nested constructions with a short-range embedded dependency, Transformers outperform RNN-LM across all conditions, with error rates close to zero. However, when the embedded dependency is long-range, their performance dramatically drops to below chance, similarly to the case of RNNs. The mere addition of a short prepositional phrase (`near the cabinet' in the example shown in figure 1) to the embedded dependency causes model performance to drop from near perfect to below chance level. Thus, contrary to what might be expected based on their much improved performance and the fact that they are trained on substantially more data, Transformer models share RNNs' shortcoming when it comes to recursive, structure-sensitive, processing.

Last, all models made more errors when trying to carry a noun in the singular across dependencies which involved a plural noun, than in the converse situation. Interestingly, this bias towards greater interference by plural than by singular is opposite to that reported in Italian RNN-LMs \citep{lakretz2020nested}, and is akin to the Markedness Effect reported for humans.

\section{Related Work}
In psycholinguistics, grammatical agreement became a standard method to probe online syntactic processing in humans \citep{Bock:Miller:1991, franck2002subject}, since it is ruled by hierarchical structures rather than by the linear order of words in a sentence. More recently, it has also become a standard way to probe grammatical generalization in NLMs \citep{Linzen:etal:2016,Bernardy:Lappin:2017,giulianelli-etal-2018-hood,Gulordava:etal:2018,jumelet-etal-2019-analysing,kersten-etal-2021-attention, Lakretz:etal:2019,sinha2021masked}, pointing to both similarities and differences between human and model error patterns. 

\citet{Lakretz:etal:2019} showed that RNN-LMs trained on a large corpus with English sentences develop a number-propagation mechanism for long-range dependencies. The core circuit of this mechanism was found to be extremely sparse, comprising of only a very small number of units. This sparsity of the mechanism suggests that models are not able to process two long-distance dependencies simultaneously, and indeed, this was later confirmed in simulations \citep{lakretz2020nested}. Inspired by this finding, \citet{lakretz2020nested} conducted a following experiment with humans, which showed that they, too, make more errors on nested long-range dependencies. However, contrary to LMs, their performance was above chance on these constructions. This finding suggests that human recursive processing remains significantly better than that of RNN-LMs.

Recursive processing of nested constructions in RNN-LMs was also studied using artificial grammars \citep{cleeremans1989finite, servan1991graded, Gers:Schmidhuber:2001, Christiansen:Chater:1999, hewitt2020rnns}. Recently, \citet{suzgun2019memory} showed that memory-augmented RNNs can capture recursive regularities of Dyck languages (also known as "bracket languages"). However, when tested on a simple extension of these languages, RNN-LMs failed to generalize to unseen data with a greater nesting depth \citep{lakretz2020recursion}. Specifically, the models failed also in cases in which the training data contained deep structures, up to five levels of nesting. This suggests that the poor recursive processing of RNN-LMs is not merely due to shallow nesting depth in natural data, which is typically not more than two \citep{karlsson2007constraints}.

Taken together, previous work suggests that RNN-LMs struggle to capture recursive regularities in either natural or artificial data. Inspired by this line of work, we focus here on Transformer LMs: do they show different patterns when it comes to processing recursive structures? Do they better approximate human ability for recursion?
\section{Experimental Setup}

We largely follow the experimental setup of \citet{lakretz2020nested}, but consider a different language (English instead of Italian) and a different set of models.

\paragraph{Data}\label{ssec:stimuli}
We consider two number-agreement tasks (\textit{NA-tasks}): \textit{Short-Nested} and \textit{Long-Nested}. 
Both tasks contain two subject-verb dependencies; they differ in terms of whether the embedded dependency is \emph{short-} or \emph{long-range}.
In \textit{short-nested}, the subject and verb in the nested dependency are adjacent (see Figure~\ref{fig:constructions}a).
They are embedded in a sentence by inserting an object-relative clause to modify the subject of a different sentence. The \textit{Long-Nested} task (Figure~\ref{fig:constructions}b) uses the same constructions, except that an additional three-word prepositional phrase (``near the cabinet'') is added in the embedded dependency. We recently made this test data set available in the BigBench collaborative benchmark.\footnote{\url{https://github.com/google/BIG-bench/tree/main/bigbench/benchmark_tasks/subject_verb_agreement}}

\paragraph{Models}\label{ssec:models}
We run experiments with all causal transformer-based NLMs that are currently compatible with the BigBench framework, available from HuggingFace.\footnote{\url{https://huggingface.co/transformers/}} 
Specifically, we include four GPT-2 models that differed in size: GPT2, GPT2-Medium, GPT2-Large and GPT-XL \citep{radford2019language}. In addition, as a baseline, we conduct an experiment with an English LSTM-LM, which was studied in numerous work in the past \citep{Gulordava:etal:2018}.

\newcommand{\mainLineWidth}{1}
\newcommand{\embedLineWidth}{2}
\newcommand{\bigDeltaY}{1}
\newcommand{\smallDeltaY}{0.5}
\newcommand{\offsetY}{0.5}

\newcommand{\objrelX}{0.1}
\newcommand{\objrelY}{0}
\newcommand{\objrelMainSubjectX}{-1.5}
\newcommand{\objrelMainVerbX}{2}
\newcommand{\objrelEmbedSubjectX}{0.5}
\newcommand{\objrelEmbedVerbX}{1.4}

\newcommand{\objrelNounppX}{0}
\newcommand{\objrelNounppY}{-6}
\newcommand{\objrelNounppMainSubjectX}{-2.9}
\newcommand{\objrelNounppMainVerbX}{3.2}
\newcommand{\objrelNounppEmbedSubjectX}{-0.9}
\newcommand{\objrelNounppEmbedVerbX}{2.5}

\begin{figure}
\centering

\begin{subfigure}{0.5\textwidth}
    \centering
     \begin{tikzpicture}
        \node[align=left] at (\objrelX,\objrelY) {The keys that the man holds are ...};
        
        \draw [line width=\mainLineWidth, black] (\objrelX+\objrelMainSubjectX,\objrelY+\offsetY) -- (\objrelX+\objrelMainSubjectX,\objrelY+\offsetY+\bigDeltaY);
        \draw [line width=\mainLineWidth, black] (\objrelX+\objrelMainSubjectX,\objrelY+\offsetY+\bigDeltaY) -- (\objrelX+\objrelMainVerbX,\objrelY+\offsetY+\bigDeltaY);
        \draw [line width=\mainLineWidth, black] (\objrelX+\objrelMainVerbX,\objrelY+\offsetY+\bigDeltaY) -- (\objrelX+\objrelMainVerbX,\objrelY+\offsetY);
    
        \draw [line width=\embedLineWidth, black] (\objrelX+\objrelEmbedSubjectX,\objrelY+\offsetY) -- (\objrelX+\objrelEmbedSubjectX,\objrelY+\offsetY+\smallDeltaY);
        \draw [line width=\embedLineWidth, black] (\objrelX+\objrelEmbedSubjectX,\objrelY+\offsetY+\smallDeltaY) -- (\objrelX+\objrelEmbedVerbX,\objrelY+\offsetY+\smallDeltaY);
        \draw [line width=\embedLineWidth, black] (\objrelX+\objrelEmbedVerbX,\objrelY+\offsetY+\smallDeltaY) -- (\objrelX+\objrelEmbedVerbX,\objrelY+\offsetY);
    \end{tikzpicture}
    \subcaption{Short-Nested}
\end{subfigure}

\begin{subfigure}{.5\textwidth}
    \centering
    \begin{tikzpicture}
        \node[align=left] at (\objrelNounppX,\objrelNounppY) {The keys that the man \underline{near the cabinet} holds are ...};
        
        \draw [line width=\mainLineWidth, black] (\objrelNounppX+\objrelNounppMainSubjectX,\objrelNounppY+\offsetY) -- (\objrelNounppX+\objrelNounppMainSubjectX,\objrelNounppY+\offsetY+\bigDeltaY);
        \draw [line width=\mainLineWidth, black] (\objrelNounppX+\objrelNounppMainSubjectX,\objrelNounppY+\offsetY+\bigDeltaY) -- (\objrelNounppX+\objrelNounppMainVerbX,\objrelNounppY+\offsetY+\bigDeltaY);
        \draw [line width=\mainLineWidth, black] (\objrelNounppX+\objrelNounppMainVerbX,\objrelNounppY+\offsetY+\bigDeltaY) -- (\objrelNounppX+\objrelNounppMainVerbX,\objrelNounppY+\offsetY);
    
        \draw [line width=\embedLineWidth, black] (\objrelNounppX+\objrelNounppEmbedSubjectX,\objrelNounppY+\offsetY) -- (\objrelNounppX+\objrelNounppEmbedSubjectX,\objrelNounppY+\offsetY+\smallDeltaY);
        \draw [line width=\embedLineWidth, black] (\objrelNounppX+\objrelNounppEmbedSubjectX,\objrelNounppY+\offsetY+\smallDeltaY) -- (\objrelNounppX+\objrelNounppEmbedVerbX,\objrelNounppY+\offsetY+\smallDeltaY);
        \draw [line width=\embedLineWidth, black] (\objrelNounppX+\objrelNounppEmbedVerbX,\objrelNounppY+\offsetY+\smallDeltaY) -- (\objrelNounppX+\objrelNounppEmbedVerbX,\objrelNounppY+\offsetY);
    \end{tikzpicture}
    \subcaption{Long-Nested}
\end{subfigure}
    \caption{Experimental Design: the two number-agreement tasks -- \textit{Short-Nested} and \textit{Long-Nested}. In Short-Nested, the embedded dependency (in bold) is short-range (in bold); in Long-Nested, it is long-range, through the insertion of a three-word prepositional phrase.}\label{fig:constructions}
\end{figure}
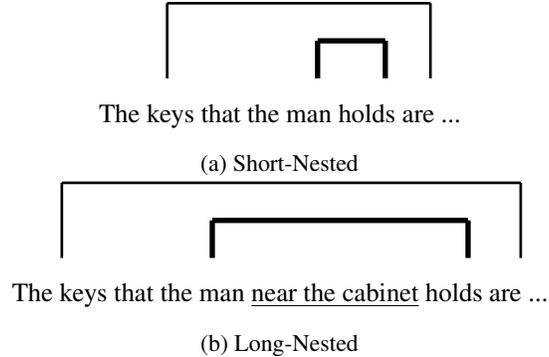

\paragraph{Model evaluation}\label{ssec:model_eval}
Following previous work, we evaluated model performance on grammatical agreement by comparing the output probabilities of the model for the correct (e.g., `are') vs. wrong (e.g., `is') verb form. 
For both tasks (short- and long-nested), we evaluated model performance on subject-verb agreement for both the embedded and the inner verb, and separately for each task condition (see SM for details).

\section{Results}

\begin{figure*}[ht]

    \centering
    \begin{subfigure}{\textwidth}
            \centering
            \includegraphics[width=0.25\linewidth]{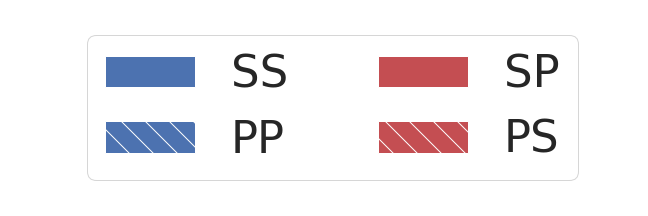}
    \end{subfigure}
    \begin{subfigure}{0.95\textwidth}
            \centering
            \includegraphics[width=\linewidth]{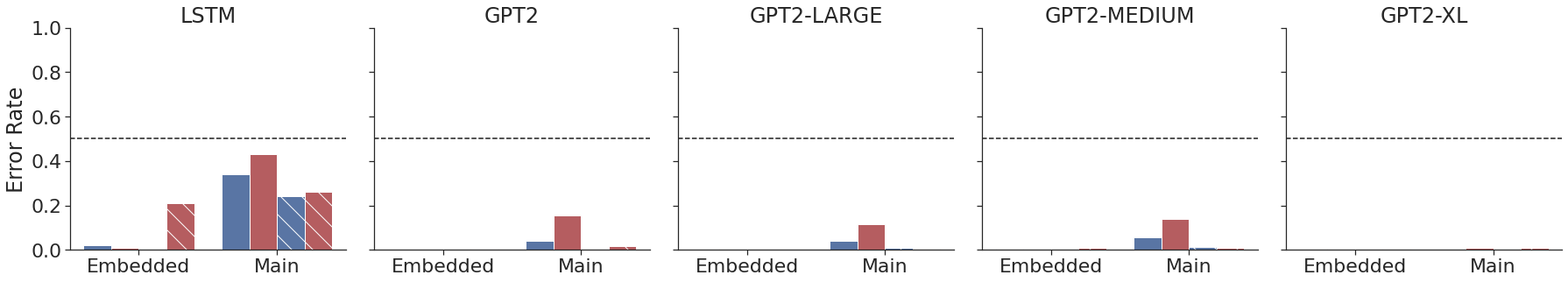}
            \subcaption{Short-Nested.}
    \end{subfigure}
    
    \bigskip
        
    \centering
    \begin{subfigure}{\textwidth}
            \centering
            \includegraphics[width=0.5\linewidth]{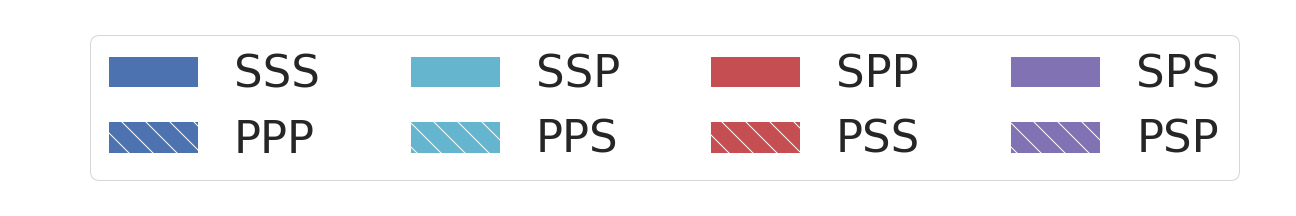}
    \end{subfigure}
    \begin{subfigure}{0.95\textwidth}
            \centering
            \includegraphics[width=\linewidth]{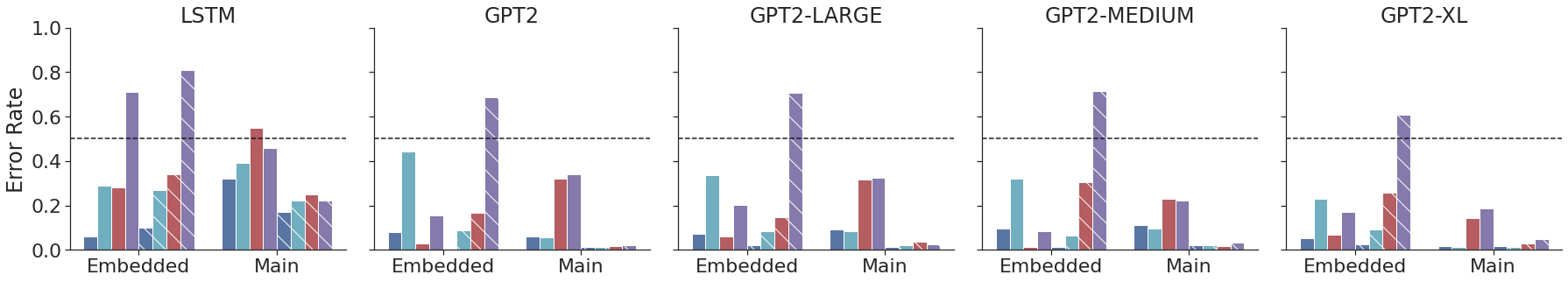}
            \subcaption{Long-Nested.}
    \end{subfigure}

    \caption{Error rates on nested constructions for all models, for both the main (`keys'-`are') and embedded (`man'-`holds') agreements. Conditions are marked by the value of the grammatical number of all nouns in the sentence. For example, condition SP in Short-Nested means that the first (main) noun is singular and the second (embedded) one is plural. While error-rates are near zero for Short-Nested, they are worse than chance-level for one of the incongruent conditions of Long-Nested, consistently across all models. In this condition (PSP), grammatical agreement is with respect to the second noun, which is singular.}
    \label{fig:english-results}
\end{figure*}
\subsection{Short-Nested task}
In Figure \ref{fig:english-results}a, we show model performance on the Short-Nested task for all models. 
Overall, the English LSTM made more errors on the main (outer) dependency compared to the embedded (inner) one, with more than 20\% errors, across all four conditions. In contrast, Transformers, and in particular GPT2-XL, achieved close to perfect performance across all conditions, on both the embedded and main dependency. For GPT2, GPT2-Medium and Large, the longer main dependency was, however, overall more difficult than the embedded one, but with no more than 20\% errors in the incongruent conditions (SP and PS; Table S2).

Interestingly, consistently across all models, both Transformers and the LSTM model made more errors on conditions in which the agreement was with respect to singular, compared to plural. 

\subsection{Long-Nested task}
In Figure \ref{fig:english-results}b, we further show the performance of all models for the Long-Nested task. Overall, all models made more errors across all conditions compared to Short-Nested, but with the same tendency of making more errors on dependencies with respect to singular compared to plural. The most striking difference between the two tasks was the performance of the models on the embedded dependency. In particular, for Transformers, their error rate was close to zero in Short-Nested, but dropped to below-chance on one of the incongurent conditions (PSP) in Long-Nested. Similarly, For the LSTM, this was the case for both incongruent cases (PSP and SPS).

In contrast to the embedded dependency, all models performed above chance on the main, longer, dependency. This shows that for Long-Nested, the length of the dependency affected model performance less than the presence of recursive embedding.
\section{Discussion}
In this study, we evaluated the recursive abilities of Transformer LMs on two number-agreement tasks that were previously shown to be exceptionally challenging for LSTM language models. Our experiments showed that, overall, Transformers outperformed LSTM-LMs, and in particular, achieved close to perfect performance on short embedded dependencies. However, similarly to LSTM-LMs, syntactic processing in the models was found to be `brittle' -- the addition of only a short prepositional phrase to the embedded dependency caused model performance to sharply drop to below chance level.

Furthermore, we found that all models showed a bias towards plural and therefore err more when the subject of a verb is in the singular. A similar bias was previously observed in Italian LSTM models \citep{lakretz2020nested}, however, in the opposite direction, with more errors on plural dependencies. We hypothesize that this difference might be due to marking of the verb form, given that in English, the marked form of the verb is singular, whereas in Italian, it is plural. Related biases were previous reported for humans in both languages, a phenomenon known as the Markedness Effect \citep{Bock:Miller:1991, vigliocco1995constructing}. The relation between emerging biases in NLMs and humans is an interesting topic for future work.

In LSTM language models, the poor performance was predicted by the underlying neural mechanism for grammatical agreement identified in the models \citep{Lakretz:etal:2019, lakretz2020nested}. 
The fact that Transformer models perform similarly poorly on these constructions, and on the same dependency (inner), raises interesting questions.
Do transformers use syntactic-processing strategies akin to those emerged in RNN-LMs?
And what does that tell us about the data that those models are trained on and about the potential processes that humans may use to process such constructions \citep{lakretz2020limits}?

However, currently, the neural mechanisms underlying syntactic processing in transformers are poorly understood \citep{belinkov2019analysis}. 
Our findings of below-chance performance by transformer models calls for a further investigation in \emph{how} these models achieve their earlier found successes on syntactic related tasks, and why they generalise so poorly on constructions which only minimally differ (a single three-word prepositional phrase) from the constructions they process well. 

\bibliographystyle{acl_natbib}
\bibliography{anthology,yair}

\begin{thebibliography}{34}
\expandafter\ifx\csname natexlab\endcsname\relax\def\natexlab#1{#1}\fi

\bibitem[{Belinkov and Glass(2019)}]{belinkov2019analysis}
Yonatan Belinkov and James Glass. 2019.
\newblock Analysis methods in neural language processing: A survey.
\newblock \emph{Transactions of the Association for Computational Linguistics},
  7:49--72.

\bibitem[{Bernardy and Lappin(2017)}]{Bernardy:Lappin:2017}
{Jean-Philippe} Bernardy and Shalom Lappin. 2017.
\newblock Using deep neural networks to learn syntactic agreement.
\newblock \emph{Linguistic Issues in Language Technology}, 15(2):1--15.

\bibitem[{Bock and Miller(1991)}]{Bock:Miller:1991}
Kathryn Bock and Carol Miller. 1991.
\newblock Broken agreement.
\newblock \emph{Cognitive Psychology}, 23(1):45--93.

\bibitem[{Chomsky(2000)}]{Chomsky:2000}
Noam Chomsky. 2000.
\newblock Minimalist inquiries: The framework.
\newblock In Roger Martin, David Michaels, Juan Uriagereka, and Samuel Keyser,
  editors, \emph{Step by step: Essays on minimalist syntax in honor of {Howard
  Lasnik}}, pages 89--155. MIT Press, Cambridge, MA.

\bibitem[{Christiansen and Chater(1999)}]{Christiansen:Chater:1999}
Morten Christiansen and Nick Chater. 1999.
\newblock Toward a connectionist model of recursion in human linguistic
  performance.
\newblock \emph{Cognitive Science}, 23(2):157--205.

\bibitem[{Cleeremans et~al.(1989)Cleeremans, Servan-Schreiber, and
  McClelland}]{cleeremans1989finite}
Axel Cleeremans, David Servan-Schreiber, and James~L McClelland. 1989.
\newblock Finite state automata and simple recurrent networks.
\newblock \emph{Neural computation}, 1(3):372--381.

\bibitem[{Dehaene et~al.(2015)Dehaene, Meyniel, Wacongne, Wang, and
  Pallier}]{Dehaene:etal:2015}
Stanislas Dehaene, Florent Meyniel, Catherine Wacongne, Liping Wang, and
  Christophe Pallier. 2015.
\newblock The neural representation of sequences: From transition probabilities
  to algebraic patterns and linguistic trees.
\newblock \emph{Neuron}, 88(1):2--19.

\bibitem[{Franck et~al.(2002)Franck, Vigliocco, and Nicol}]{franck2002subject}
Julie Franck, Gabriella Vigliocco, and Janet Nicol. 2002.
\newblock Subject-verb agreement errors in french and english: The role of
  syntactic hierarchy.
\newblock \emph{Language and cognitive processes}, 17(4):371--404.

\bibitem[{Gers and Schmidhuber(2001)}]{Gers:Schmidhuber:2001}
Felix Gers and J\"{u}rgen Schmidhuber. 2001.
\newblock {LSTM} recurrent networks learn simple context-free and
  context-sensitive languages.
\newblock \emph{IEEE Transactions on Neural Networks}, 12(6):1333--1340.

\bibitem[{Giulianelli et~al.(2018)Giulianelli, Harding, Mohnert, Hupkes, and
  Zuidema}]{giulianelli-etal-2018-hood}
Mario Giulianelli, Jack Harding, Florian Mohnert, Dieuwke Hupkes, and Willem
  Zuidema. 2018.
\newblock \href {https://doi.org/10.18653/v1/W18-5426} {Under the hood: Using
  diagnostic classifiers to investigate and improve how language models track
  agreement information}.
\newblock In \emph{Proceedings of the 2018 {EMNLP} Workshop {B}lackbox{NLP}:
  Analyzing and Interpreting Neural Networks for {NLP}}, pages 240--248,
  Brussels, Belgium. Association for Computational Linguistics.

\bibitem[{Goldberg(2019)}]{goldberg2019assessing}
Yoav Goldberg. 2019.
\newblock Assessing bert's syntactic abilities.
\newblock \emph{arXiv preprint arXiv:1901.05287}.

\bibitem[{Gulordava et~al.(2018)Gulordava, Bojanowski, Grave, Linzen, and
  Baroni}]{Gulordava:etal:2018}
Kristina Gulordava, Piotr Bojanowski, Edouard Grave, Tal Linzen, and Marco
  Baroni. 2018.
\newblock Colorless green recurrent networks dream hierarchically.
\newblock In \emph{Proceedings of NAACL}, pages 1195--1205, New Orleans, LA.

\bibitem[{Hauser et~al.(2002)Hauser, Chomsky, and Fitch}]{Hauser:etal:2002}
Marc Hauser, Noam Chomsky, and Tecumseh Fitch. 2002.
\newblock The faculty of language: What is it, who has it, and how did it
  evolve?
\newblock \emph{Science}, 298(5598):1569--1579.

\bibitem[{Hewitt et~al.(2020)Hewitt, Hahn, Ganguli, Liang, and
  Manning}]{hewitt2020rnns}
John Hewitt, Michael Hahn, Surya Ganguli, Percy Liang, and Christopher~D
  Manning. 2020.
\newblock Rnns can generate bounded hierarchical languages with optimal memory.
\newblock \emph{arXiv preprint arXiv:2010.07515}.

\bibitem[{Hochreiter and Schmidhuber(1997)}]{Hochreiter:Schmidhuber:1997}
Sepp Hochreiter and J\"{u}rgen Schmidhuber. 1997.
\newblock Long short-term memory.
\newblock \emph{Neural Computation}, 9(8):1735--1780.

\bibitem[{Hu et~al.(2020)Hu, Gauthier, Qian, Wilcox, and
  Levy}]{hu2020systematic}
Jennifer Hu, Jon Gauthier, Peng Qian, Ethan Wilcox, and Roger~P Levy. 2020.
\newblock A systematic assessment of syntactic generalization in neural
  language models.
\newblock \emph{arXiv preprint arXiv:2005.03692}.

\bibitem[{Jumelet et~al.(2021)Jumelet, Denic, Szymanik, Hupkes, and
  Steinert-Threlkeld}]{jumelet-etal-2021-language}
Jaap Jumelet, Milica Denic, Jakub Szymanik, Dieuwke Hupkes, and Shane
  Steinert-Threlkeld. 2021.
\newblock \href {https://doi.org/10.18653/v1/2021.findings-acl.439} {Language
  models use monotonicity to assess {NPI} licensing}.
\newblock In \emph{Findings of the Association for Computational Linguistics:
  ACL-IJCNLP 2021}, pages 4958--4969, Online. Association for Computational
  Linguistics.

\bibitem[{Jumelet et~al.(2019)Jumelet, Zuidema, and
  Hupkes}]{jumelet-etal-2019-analysing}
Jaap Jumelet, Willem Zuidema, and Dieuwke Hupkes. 2019.
\newblock \href {https://doi.org/10.18653/v1/K19-1001} {Analysing neural
  language models: Contextual decomposition reveals default reasoning in number
  and gender assignment}.
\newblock In \emph{Proceedings of the 23rd Conference on Computational Natural
  Language Learning (CoNLL)}, pages 1--11, Hong Kong, China. Association for
  Computational Linguistics.

\bibitem[{Karlsson(2007)}]{karlsson2007constraints}
Fred Karlsson. 2007.
\newblock Constraints on multiple center-embedding of clauses.
\newblock \emph{Journal of Linguistics}, 43(2):365--392.

\bibitem[{Kersten et~al.(2021)Kersten, Wong, Jumelet, and
  Hupkes}]{kersten-etal-2021-attention}
Tom Kersten, Hugh~Mee Wong, Jaap Jumelet, and Dieuwke Hupkes. 2021.
\newblock \href {https://doi.org/10.18653/v1/2021.deelio-1.13} {Attention vs
  non-attention for a shapley-based explanation method}.
\newblock In \emph{Proceedings of Deep Learning Inside Out (DeeLIO): The 2nd
  Workshop on Knowledge Extraction and Integration for Deep Learning
  Architectures}, pages 129--139, Online. Association for Computational
  Linguistics.

\bibitem[{Lakretz et~al.(2020)Lakretz, Dehaene, and King}]{lakretz2020limits}
Yair Lakretz, Stanislas Dehaene, and Jean-R{\'e}mi King. 2020.
\newblock What limits our capacity to process nested long-range dependencies in
  sentence comprehension?
\newblock \emph{Entropy}, 22(4):446.

\bibitem[{Lakretz et~al.(2021{\natexlab{a}})Lakretz, Desbordes, King,
  Crabb{\'e}, Oquab, and Dehaene}]{lakretz2020recursion}
Yair Lakretz, Th{\'e}o Desbordes, Jean-R{\'e}mi King, Beno{\^\i}t Crabb{\'e},
  Maxime Oquab, and Stanislas Dehaene. 2021{\natexlab{a}}.
\newblock Can rnns learn recursive nested subject-verb agreements?
\newblock \emph{arXiv preprint arXiv:2101.02258}.

\bibitem[{Lakretz et~al.(2021{\natexlab{b}})Lakretz, Hupkes, Vergallito,
  Marelli, Baroni, and Dehaene}]{lakretz2020nested}
Yair Lakretz, Dieuwke Hupkes, Alessandra Vergallito, Marco Marelli, Marco
  Baroni, and Stanislas Dehaene. 2021{\natexlab{b}}.
\newblock Mechanisms for handling nested dependencies in neural-network
  language models and humans.
\newblock \emph{Cognition}, page 104699.

\bibitem[{Lakretz et~al.(2019)Lakretz, Kruszewski, Desbordes, Hupkes, Dehaene,
  and Baroni}]{Lakretz:etal:2019}
Yair Lakretz, Germ\'{a}n Kruszewski, Theo Desbordes, Dieuwke Hupkes, Stanislas
  Dehaene, and Marco Baroni. 2019.
\newblock The emergence of number and syntax units in {LSTM} language models.
\newblock In \emph{Proceedings of NAACL}, pages 11--20, Minneapolis, MN.

\bibitem[{Linzen et~al.(2016)Linzen, Dupoux, and Goldberg}]{Linzen:etal:2016}
Tal Linzen, Emmanuel Dupoux, and Yoav Goldberg. 2016.
\newblock Assessing the ability of {LSTM}s to learn syntax-sensitive
  dependencies.
\newblock \emph{Transactions of the Association for Computational Linguistics},
  4:521--535.

\bibitem[{Marvin and Linzen(2018)}]{marvin-linzen-2018-targeted}
Rebecca Marvin and Tal Linzen. 2018.
\newblock \href {https://doi.org/10.18653/v1/D18-1151} {Targeted syntactic
  evaluation of language models}.
\newblock In \emph{Proceedings of the 2018 Conference on Empirical Methods in
  Natural Language Processing}, pages 1192--1202, Brussels, Belgium.
  Association for Computational Linguistics.

\bibitem[{Petty and Frank(2021)}]{petty2021transformers}
Jackson Petty and Robert Frank. 2021.
\newblock Transformers generalize linearly.
\newblock \emph{arXiv preprint arXiv:2109.12036}.

\bibitem[{Radford et~al.(2019)Radford, Wu, Child, Luan, Amodei, Sutskever
  et~al.}]{radford2019language}
Alec Radford, Jeffrey Wu, Rewon Child, David Luan, Dario Amodei, Ilya
  Sutskever, et~al. 2019.
\newblock Language models are unsupervised multitask learners.
\newblock \emph{OpenAI blog}, 1(8):9.

\bibitem[{Servan-Schreiber et~al.(1991)Servan-Schreiber, Cleeremans, and
  McClelland}]{servan1991graded}
David Servan-Schreiber, Axel Cleeremans, and James~L McClelland. 1991.
\newblock Graded state machines: The representation of temporal contingencies
  in simple recurrent networks.
\newblock \emph{Machine Learning}, 7(2-3):161--193.

\bibitem[{Sinha et~al.(2021)Sinha, Jia, Hupkes, Pineau, Williams, and
  Kiela}]{sinha2021masked}
Koustuv Sinha, Robin Jia, Dieuwke Hupkes, Joelle Pineau, Adina Williams, and
  Douwe Kiela. 2021.
\newblock \href {https://arxiv.org/abs/2104.06644} {Masked language modeling
  and the distributional hypothesis: Order word matters pre-training for
  little}.
\newblock \emph{CoRR}, abs/2104.06644.

\bibitem[{Suzgun et~al.(2019)Suzgun, Gehrmann, Belinkov, and
  Shieber}]{suzgun2019memory}
Mirac Suzgun, Sebastian Gehrmann, Yonatan Belinkov, and Stuart~M Shieber. 2019.
\newblock Memory-augmented recurrent neural networks can learn generalized dyck
  languages.
\newblock \emph{arXiv preprint arXiv:1911.03329}.

\bibitem[{Vaswani et~al.(2017)Vaswani, Shazeer, Parmar, Uszkoreit, Jones,
  Gomez, Kaiser, and Polosukhin}]{vaswani2017attention}
Ashish Vaswani, Noam Shazeer, Niki Parmar, Jakob Uszkoreit, Llion Jones,
  Aidan~N Gomez, {\L}ukasz Kaiser, and Illia Polosukhin. 2017.
\newblock \href
  {https://papers.nips.cc/paper/2017/file/3f5ee243547dee91fbd053c1c4a845aa-Paper.pdf}
  {Attention is all you need}.
\newblock In \emph{Advances in Neural Information Processing Systems}, pages
  5998--6008.

\bibitem[{Vigliocco et~al.(1995)Vigliocco, Butterworth, and
  Semenza}]{vigliocco1995constructing}
Gabriella Vigliocco, Brian Butterworth, and Carlo Semenza. 1995.
\newblock Constructing subject-verb agreement in speech: The role of semantic
  and morphological factors.
\newblock \emph{Journal of Memory and Language}, 34(2):186--215.

\bibitem[{Warstadt et~al.(2020)Warstadt, Parrish, Liu, Mohananey, Peng, Wang,
  and Bowman}]{warstadt-etal-2020-blimp}
Alex Warstadt, Alicia Parrish, Haokun Liu, Anhad Mohananey, Wei Peng, Sheng-Fu
  Wang, and Samuel~R. Bowman. 2020.
\newblock \href {https://aclanthology.org/2020.scil-1.47} {{BL}i{MP}: A
  benchmark of linguistic minimal pairs for {E}nglish}.
\newblock In \emph{Proceedings of the Society for Computation in Linguistics
  2020}, pages 409--410, New York, New York. Association for Computational
  Linguistics.

\end{thebibliography}


\end{document}



\begin{table}[ph]
    \setlength\tabcolsep{3mm}
\small
\centering
\begin{tabular}{lll}
\multicolumn{3}{c}{\centering \textit{Number-Agreement tasks}}\\
\hline
\emph{Short-Nested} & & \texttt{NP$_a$ that NP$_b$ V$_b$ V$_a$ } \\
\hline
& SS & The \textbf{key} that the \emph{man} \emph{holds} \textbf{is} \\
& SP & The \textbf{key} that the \emph{men} \emph{hold} \textbf{is} \\
& PS & The \textbf{keys} that the \emph{men} \emph{holds} \textbf{are} \\
& PP & The \textbf{keys} that the \emph{man} \emph{holds} \textbf{are} \\
\\
\hline
\emph{Long-Nested} & & \texttt{NP$_a$ that NP$_b$ P NP$_c$ V$_b$ V$_a$}\\
\hline
& SSS & The \textbf{key} that the \emph{man} near the \underline{cabinet} \emph{holds} \textbf{is} \\
& SSP & The \textbf{key} that the \emph{man} near the \underline{cabinets} \emph{holds} \textbf{is} \\
& SPS & The \textbf{key} that the \emph{men} near the \underline{cabinet} \emph{hold} \textbf{is} \\
& SPP & The \textbf{key} that the \emph{men} near the \underline{cabinets} \emph{hold} \textbf{is} \\
& PSS & The \textbf{keys} that the \emph{man} near the \underline{cabinet} \emph{holds} \textbf{are} \\
& PSP & The \textbf{keys} that the \emph{man} near the \underline{cabinets} \emph{holds} \textbf{are} \\
& PPS & The \textbf{keys} that the \emph{men} near the \underline{cabinet} \emph{hold} \textbf{are} \\
& PPP & The \textbf{keys} that the \emph{men} near the \underline{cabinet} \emph{hold} \textbf{are} \\
\hline
\end{tabular}
\caption{\textbf{The Short- and Long-Nested Number-Agreement tasks.}
The first column denotes the name of the task, the second shows the conditions for each task, the third shows the sentence template, where \texttt{NP} is used as an abbreviation of \texttt{Det N}.
The indices $a$, $b$ mark the subject-verb dependencies in the templates. 
For example, in \emph{Long-Nested}, there are three nouns and two verbs, the indices $a$ and $b$ indicate that the last verb \texttt{V$_a$} is syntactically dependent on the first noun phrase \texttt{NP$_a$}, whereas the penultimate verb \texttt{V$_b$} instead should match the features of the second noun phrase \texttt{NP$_b$}. Below each template, and example for each condition is given. Bold and italic face highlight the dependencies marked by the indices in the templates. For each agreement task, we systematically vary the \emph{number} of all nouns in the template, resulting in four different conditions (SS, SP, PS and PP) for the \emph{Short-Nested} and eight different conditions (SSS, SSP, SPS, SPP, PSS, PSP, PPS and PPP) for {Long-Nested}.
\label{tab:na-tasks-overview}}
\end{table}

\begin{tabular}{llllr}
\toprule
Model &      NA-Task & Dependency & Condition &  Error Rate \\
\midrule
 LSTM & Short-Nested &       Main &        SS &        0.34 \\
 LSTM & Short-Nested &       Main &        SP &        0.43 \\
 LSTM & Short-Nested &       Main &        PS &        0.26 \\
 LSTM & Short-Nested &       Main &        PP &        0.24 \\
 LSTM & Short-Nested &   Embedded &        SS &        0.02 \\
 LSTM & Short-Nested &   Embedded &        SP &        0.01 \\
 LSTM & Short-Nested &   Embedded &        PS &        0.21 \\
 LSTM & Short-Nested &   Embedded &        PP &        0.00 \\
 LSTM &  Long-Nested &       Main &       SSS &        0.32 \\
 LSTM &  Long-Nested &       Main &       SSP &        0.39 \\
 LSTM &  Long-Nested &       Main &       SPS &        0.46 \\
 LSTM &  Long-Nested &       Main &       SPP &        0.55 \\
 LSTM &  Long-Nested &       Main &       PSS &        0.25 \\
 LSTM &  Long-Nested &       Main &       PSP &        0.22 \\
 LSTM &  Long-Nested &       Main &       PPS &        0.22 \\
 LSTM &  Long-Nested &       Main &       PPP &        0.17 \\
 LSTM &  Long-Nested &   Embedded &       SSS &        0.06 \\
 LSTM &  Long-Nested &   Embedded &       SSP &        0.29 \\
 LSTM &  Long-Nested &   Embedded &       SPS &        0.71 \\
 LSTM &  Long-Nested &   Embedded &       SPP &        0.28 \\
 LSTM &  Long-Nested &   Embedded &       PSS &        0.34 \\
 LSTM &  Long-Nested &   Embedded &       PSP &        0.81 \\
 LSTM &  Long-Nested &   Embedded &       PPS &        0.27 \\
 LSTM &  Long-Nested &   Embedded &       PPP &        0.10 \\
\bottomrule
\end{tabular}

\begin{tabular}{llllr}
\toprule
Model &      NA-Task & Dependency & Condition &  Error Rate \\
\midrule
 GPT2 & Short-Nested &       Main &        SS &        0.04 \\
 GPT2 & Short-Nested &       Main &        SP &        0.16 \\
 GPT2 & Short-Nested &       Main &        PS &        0.02 \\
 GPT2 & Short-Nested &       Main &        PP &        0.00 \\
 GPT2 & Short-Nested &   Embedded &        SS &        0.00 \\
 GPT2 & Short-Nested &   Embedded &        SP &        0.00 \\
 GPT2 & Short-Nested &   Embedded &        PS &        0.00 \\
 GPT2 & Short-Nested &   Embedded &        PP &        0.00 \\
 GPT2 &  Long-Nested &       Main &       SSS &        0.06 \\
 GPT2 &  Long-Nested &       Main &       SSP &        0.05 \\
 GPT2 &  Long-Nested &       Main &       SPS &        0.34 \\
 GPT2 &  Long-Nested &       Main &       SPP &        0.32 \\
 GPT2 &  Long-Nested &       Main &       PSS &        0.02 \\
 GPT2 &  Long-Nested &       Main &       PSP &        0.02 \\
 GPT2 &  Long-Nested &       Main &       PPS &        0.01 \\
 GPT2 &  Long-Nested &       Main &       PPP &        0.01 \\
 GPT2 &  Long-Nested &   Embedded &       SSS &        0.08 \\
 GPT2 &  Long-Nested &   Embedded &       SSP &        0.44 \\
 GPT2 &  Long-Nested &   Embedded &       SPS &        0.16 \\
 GPT2 &  Long-Nested &   Embedded &       SPP &        0.03 \\
 GPT2 &  Long-Nested &   Embedded &       PSS &        0.17 \\
 GPT2 &  Long-Nested &   Embedded &       PSP &        0.69 \\
 GPT2 &  Long-Nested &   Embedded &       PPS &        0.09 \\
 GPT2 &  Long-Nested &   Embedded &       PPP &        0.00 \\
\bottomrule
\end{tabular}

\begin{tabular}{llllr}
\toprule
      Model &      NA-Task & Dependency & Condition &  Error Rate \\
\midrule
GPT2-MEDIUM & Short-Nested &       Main &        SS &        0.06 \\
GPT2-MEDIUM & Short-Nested &       Main &        SP &        0.14 \\
GPT2-MEDIUM & Short-Nested &       Main &        PS &        0.01 \\
GPT2-MEDIUM & Short-Nested &       Main &        PP &        0.01 \\
GPT2-MEDIUM & Short-Nested &   Embedded &        SS &        0.01 \\
GPT2-MEDIUM & Short-Nested &   Embedded &        SP &        0.00 \\
GPT2-MEDIUM & Short-Nested &   Embedded &        PS &        0.01 \\
GPT2-MEDIUM & Short-Nested &   Embedded &        PP &        0.00 \\
GPT2-MEDIUM &  Long-Nested &       Main &       SSS &        0.11 \\
GPT2-MEDIUM &  Long-Nested &       Main &       SSP &        0.10 \\
GPT2-MEDIUM &  Long-Nested &       Main &       SPS &        0.22 \\
GPT2-MEDIUM &  Long-Nested &       Main &       SPP &        0.23 \\
GPT2-MEDIUM &  Long-Nested &       Main &       PSS &        0.02 \\
GPT2-MEDIUM &  Long-Nested &       Main &       PSP &        0.03 \\
GPT2-MEDIUM &  Long-Nested &       Main &       PPS &        0.02 \\
GPT2-MEDIUM &  Long-Nested &       Main &       PPP &        0.02 \\
GPT2-MEDIUM &  Long-Nested &   Embedded &       SSS &        0.10 \\
GPT2-MEDIUM &  Long-Nested &   Embedded &       SSP &        0.32 \\
GPT2-MEDIUM &  Long-Nested &   Embedded &       SPS &        0.08 \\
GPT2-MEDIUM &  Long-Nested &   Embedded &       SPP &        0.01 \\
GPT2-MEDIUM &  Long-Nested &   Embedded &       PSS &        0.30 \\
GPT2-MEDIUM &  Long-Nested &   Embedded &       PSP &        0.71 \\
GPT2-MEDIUM &  Long-Nested &   Embedded &       PPS &        0.06 \\
GPT2-MEDIUM &  Long-Nested &   Embedded &       PPP &        0.01 \\
\bottomrule
\end{tabular}

\begin{tabular}{llllr}
\toprule
     Model &      NA-Task & Dependency & Condition &  Error Rate \\
\midrule
GPT2-LARGE & Short-Nested &       Main &        SS &        0.04 \\
GPT2-LARGE & Short-Nested &       Main &        SP &        0.11 \\
GPT2-LARGE & Short-Nested &       Main &        PS &        0.01 \\
GPT2-LARGE & Short-Nested &       Main &        PP &        0.01 \\
GPT2-LARGE & Short-Nested &   Embedded &        SS &        0.00 \\
GPT2-LARGE & Short-Nested &   Embedded &        SP &        0.01 \\
GPT2-LARGE & Short-Nested &   Embedded &        PS &        0.00 \\
GPT2-LARGE & Short-Nested &   Embedded &        PP &        0.00 \\
GPT2-LARGE &  Long-Nested &       Main &       SSS &        0.09 \\
GPT2-LARGE &  Long-Nested &       Main &       SSP &        0.09 \\
GPT2-LARGE &  Long-Nested &       Main &       SPS &        0.32 \\
GPT2-LARGE &  Long-Nested &       Main &       SPP &        0.32 \\
GPT2-LARGE &  Long-Nested &       Main &       PSS &        0.04 \\
GPT2-LARGE &  Long-Nested &       Main &       PSP &        0.03 \\
GPT2-LARGE &  Long-Nested &       Main &       PPS &        0.02 \\
GPT2-LARGE &  Long-Nested &       Main &       PPP &        0.01 \\
GPT2-LARGE &  Long-Nested &   Embedded &       SSS &        0.07 \\
GPT2-LARGE &  Long-Nested &   Embedded &       SSP &        0.34 \\
GPT2-LARGE &  Long-Nested &   Embedded &       SPS &        0.20 \\
GPT2-LARGE &  Long-Nested &   Embedded &       SPP &        0.06 \\
GPT2-LARGE &  Long-Nested &   Embedded &       PSS &        0.15 \\
GPT2-LARGE &  Long-Nested &   Embedded &       PSP &        0.71 \\
GPT2-LARGE &  Long-Nested &   Embedded &       PPS &        0.08 \\
GPT2-LARGE &  Long-Nested &   Embedded &       PPP &        0.02 \\
\bottomrule
\end{tabular}

\begin{tabular}{llllr}
\toprule
  Model &      NA-Task & Dependency & Condition &  Error Rate \\
\midrule
GPT2-XL & Short-Nested &       Main &        SS &        0.00 \\
GPT2-XL & Short-Nested &       Main &        SP &        0.01 \\
GPT2-XL & Short-Nested &       Main &        PS &        0.01 \\
GPT2-XL & Short-Nested &       Main &        PP &        0.01 \\
GPT2-XL & Short-Nested &   Embedded &        SS &        0.00 \\
GPT2-XL & Short-Nested &   Embedded &        SP &        0.01 \\
GPT2-XL & Short-Nested &   Embedded &        PS &        0.00 \\
GPT2-XL & Short-Nested &   Embedded &        PP &        0.00 \\
GPT2-XL &  Long-Nested &       Main &       SSS &        0.02 \\
GPT2-XL &  Long-Nested &       Main &       SSP &        0.01 \\
GPT2-XL &  Long-Nested &       Main &       SPS &        0.19 \\
GPT2-XL &  Long-Nested &       Main &       SPP &        0.14 \\
GPT2-XL &  Long-Nested &       Main &       PSS &        0.03 \\
GPT2-XL &  Long-Nested &       Main &       PSP &        0.05 \\
GPT2-XL &  Long-Nested &       Main &       PPS &        0.01 \\
GPT2-XL &  Long-Nested &       Main &       PPP &        0.02 \\
GPT2-XL &  Long-Nested &   Embedded &       SSS &        0.05 \\
GPT2-XL &  Long-Nested &   Embedded &       SSP &        0.23 \\
GPT2-XL &  Long-Nested &   Embedded &       SPS &        0.17 \\
GPT2-XL &  Long-Nested &   Embedded &       SPP &        0.07 \\
GPT2-XL &  Long-Nested &   Embedded &       PSS &        0.26 \\
GPT2-XL &  Long-Nested &   Embedded &       PSP &        0.61 \\
GPT2-XL &  Long-Nested &   Embedded &       PPS &        0.09 \\
GPT2-XL &  Long-Nested &   Embedded &       PPP &        0.03 \\
\bottomrule
\end{tabular}